%% file: KENN_relational.tex
\documentclass{article}

\usepackage[preprint]{neurips_2022}

\usepackage[utf8]{inputenc} 
\usepackage[T1]{fontenc}    
\usepackage{hyperref}       
\usepackage{url}            
\usepackage{booktabs}       
\usepackage{amsfonts}       
\usepackage{nicefrac}       
\usepackage{microtype}      
\usepackage{xcolor}         

\usepackage{natbib}
\setcitestyle{numbers, comma}

\usepackage{amssymb}
\usepackage{amsfonts}
\usepackage{listings}

\usepackage{physics}
\usepackage{bm}
\usepackage{float}
\usepackage{scalerel}
\usepackage{tikz}
\usetikzlibrary{positioning}
\usetikzlibrary{arrows}
\usetikzlibrary{arrows.meta}
\usetikzlibrary{fit}
\usetikzlibrary{backgrounds}
\usetikzlibrary{matrix}
\usetikzlibrary{shapes.multipart}
\usetikzlibrary{decorations.pathreplacing}
\usetikzlibrary{calc}

\def\kenn{KENN}

\def\Reals{\mathbb{R}}

\def\bdelta{\bm\delta}
\def\clauses{\mathcal{K}}
\def\L{\mathcal{L}}

\def\C{\mathcal{C}}
\def\P{\mathcal{P}}

\DeclareMathOperator*{\argmax}{\arg\!max}

\def\bx{\mathbf{x}}
\def\by{\mathbf{y}}
\def\bz{\mathbf{z}}

\newcommand{\new}[1]{#1} 

\title{Knowledge Enhanced Neural Networks for relational domains}

\author{%
  Alessandro Daniele \\
  Data and Knowledge Managemen Research Unit\\
  Fondazione Bruno Kessler\\
  Trento, Italy\\
  \texttt{daniele@fbk.eu} \\
  \And
  Luciano Serafini \\
  Data and Knowledge Managemen Research Unit\\
  Fondazione Bruno Kessler\\
  Trento, Italy\\
  \texttt{serafini@fbk.eu} \\
}

\begin{document}

\maketitle

\begin{abstract}
  In the recent past, there has been a growing interest in Neural-Symbolic Integration frameworks, i.e., hybrid systems that integrate connectionist and symbolic approaches to obtain the best of both worlds. \new{In this work we focus on a specific method,} KENN
  (Knowledge Enhanced Neural Networks), a Neural-Symbolic architecture that injects prior logical knowledge into a neural network by adding on its top a residual layer that modifies the initial predictions accordingly to the knowledge. Among the advantages of this strategy, there is the inclusion of clause weights, learnable parameters that represent the strength of the clauses, meaning that the model can learn the impact of each rule on the final predictions. 
  As a special case, if the training data contradicts a constraint, \kenn\ learns to ignore it, making the system robust to the presence of wrong knowledge. In this paper, we propose an extension of \kenn\ for relational data. \new{
  One of the main advantages of \kenn\ resides in its scalability, thanks to a flexible treatment of dependencies between the rules obtained by stacking multiple logical layers. We show experimentally the efficacy of this strategy.
  The results show that \kenn\ is capable of increasing the performances of the underlying neural network, obtaining better or comparable accuracies in respect to other two related methods that combine learning with logic, requiring significantly less time for learning.}

  

\end{abstract}

\input{introduction}
\input{state_of_the_art}

\input{kenn}

\input{relational}

\input{collective}
\bibliography{KENN_relational}
\newpage

\end{document}


\input{appendix}

%% file: introduction.tex
\section{Introduction}
\label{sec:introduction}

In the last decade, deep learning approaches gained a lot of interest in the AI community, becoming the state of the art on many fields, such as Computer Vision~\cite{image_classification}, Machine Translation~\cite{machine_translation}, Speech Recognition~\cite{speech_recognition}, etc. 
\new{
Indeed, Neural networks (NNs) are suited for pattern recognition, even in the presence of noisy data. They are particularly good at mapping low-level perceptions to more abstract concepts (for instance, going from images to classes). However, it is hard for a NN to reason with these high-level abstractions. Furthermore, NNs are demanding in terms of training data. On the other hand, pure logical approaches are not suited for learning from low-level features and they struggle in the presence of noise. Nevertheless, they perform well in reasoning with highly abstract concepts and learning from a small number of samples. Given these opposite strengths and weaknesses, it is not a surprise that a lot of interest has been drawn toward Neural-Symbolic (NeSy) systems. Indeed, the goal is to combine these two paradigms to obtain the best of the two worlds.}


\new{Among NeSy methods} there is \kenn\ (\emph{Knowledge Enhanced Neural Network})~\cite{kenn}, \new{a model composed of a Neural Network enhanced with additional layers which codify logical knowledge.} \kenn\ has multiple advantages over other NeSy methods, such as its capacity to learn \emph{clause weights} and the ability to impose the knowledge not only during training but even at inference time.
In particular, \kenn\ showed remarkable results on the Predicate Detection task of Visual Relationship Detection Dataset (VRD Dataset)~\cite{visual2} using a manually curated prior knowledge proposed by~\cite{IvanThesis}, outperforming the previous state of the art results, with really good performances on the \emph{Zero Shot Learning} sub-task~\cite{kenn}. Moreover, it outperformed Logic Tensor Networks~\cite{LTN}, one of its major competitors, using the same knowledge.

Despite its good empirical results, \kenn \ has been applied only on multi-label classification tasks with no relational data. Indeed, a limitation of \kenn\ resides in its inability to take into account binary predicates. This is because \kenn \ expects the NN's predictions to be stored in a matrix format, where the columns represent different unary predicates and the rows their possible groundings (i.e., substitutions of the free variable for such predicates). 
For this reason, 
it is not straightforward to apply \kenn\ to relational data, where binary predicates are available.

In this paper,
we propose an updated version of \kenn\ which can deal with relational data.
\new{Particular attention was paid to defining a scalable strategy to deal with binary predicates, obtaining good performances in terms of execution time. Indeed, \kenn\ assumes independence between the logical rules, allowing for scalable inclusion of the underlying knowledge. However, the assumption is often violated in real scenarios, in particular in the contexts of relational domains. To deal with this problem, we propose a strategy that consists of adding multiple logical layers inside the model. We provide proof of the efficacy of this strategy in a simple scenario with two logical rules. Additionally, we tested this idea on Citeseer, a dataset for Collective Classification~\cite{collective_classification}, showing that the additional layers improve the performance of the model.
Moreover, 
}
the experiments on this dataset 
provide
a comparison between \kenn\ and two other approaches: Semantic Based Regularization (SBR)~\cite{SBR} and Relational Neural Machines (RNM)~\cite{RNM}. 

%% file: state_of_the_art.tex
\section{Related works}
\label{ch:rel_works}

Many previous works attempt to combine learning models with logical knowledge. Among them there is 
Statistical Relational Learning (SRL), a subfield of Machine Learning that aims at applying statistical methods in domains that exhibit both uncertainty and relational structure \cite{introduction_SRL}. \new{Generally speaking, SRL deals with the knowledge either by combining logic rules with probabilistic graphical models (e.g., Markov Logic Networks~\cite{MLN}, and \emph{Probabilistic Soft Logic} (PSL)~\cite{PSL}) or by extending logic programming languages to handle uncertainty (e.g., ProbLog \cite{problog}).}




The recent achievements of deep learning methods lead to a renewed interest in another line of research, called Neural-Symbolic Integration, which focuses on combining neural network architectures with logical knowledge~\cite{neural_symbolic}.
This can be achieved in multiple ways depending on the role of the knowledge. For instance, works like 
TensorLog~\cite{tensorlog}, Neural Theorem Prover (NTP)~\cite{NTP1,NTP2}, DeepProbLog~\cite{deepproblog}, Neural Logic Machines~\cite{NLM}, and NeuralLog~\cite{NeuralLog}
focus on the development of differentiable approaches for reasoning, which can be used in combination with neural networks. Another line of research comes from methods like 
$\partial$ILP~\cite{dilp}, \cite{rule_induction}, and Neural Logic Rule Layer (NLRL)~\cite{NLRL}. In these cases, the goal is to learn general knowledge from the data, either from scratch or by refining an initial knowledge. Finally, more related to our purposes, some methods focus on learning in the presence of prior knowledge, which acts as additional supervision. In this section, we are going to focus on these types of methods, since \kenn\ falls in this category.


There are mainly two approaches for learning in the presence of prior knowledge: the first consists of treating logical rules as constraints on the predictions of the neural network. The problem is reduced to maximize the satisfiability of the constraints and can be efficiently tackled by adding a regularization term in the Loss function which penalizes the unsatisfaction of the constraints. The second approach is to modify the neural network by injecting the knowledge into its structure.



Two notable examples of regularization approaches are \emph{Logic Tensor Network} (LTN)~\cite{LTN} and \emph{Semantic Based Regularization} (SBR)~\cite{SBR}. Both methods maximize the satisfaction of the constraints, expressed as FOL formulas, under a fuzzy logic semantic. 
A similar strategy is employed also by \emph{Semantic Loss Function}~\cite{semantic_loss}, but instead of relying on fuzzy logic, it optimizes the probability of the rules being true. Nevertheless, this approach is restricted to propositional logic. \cite{dl2} introduces DL2.
Nonetheless, it can be used only in the context of regression tasks, where the predicates correspond to comparison constraints (e.g. $=$, $\neq$, $\leq$).
MultiplexNet~\cite{MultiplexNet} also uses comparison constraints, but it is not restricted to them. In this case, the interest is in safety-critical domains where complete satisfaction of the constraints is desirable and the knowledge is expressed in disjunctive normal form (DNF).
\cite{adversarially} also proposes a method that regularizes the Loss, but they focus on a specific task of Natural Language Processing. Their approach differs from the others because it makes use of adversarial examples to calculate the regularization term.
Finally, in \cite{semi_supervised} 
the regularization term is calculated from the unlabelled data.

In~\cite{harnessing}, a distillation mechanism is used to inject FOL rules.
While this approach seems different from the previous ones, we argue that it can be also considered as a method based on regularization, since a teacher network (which encodes the rules) is used to regularize the Loss applied to a student network.
Finally, in~\cite{old_dog}, a new technique is defined which adds the knowledge at the level of the labels by modifying them after each training step. 


Model-based approaches work pretty differently from the methods seen so far. The main difference lies in the way logic formulas are used:
the knowledge becomes part of the classifier.
Instead, approaches based on regularization force the constraints satisfaction solely at training time. As a consequence, there are no guarantees that they will be satisfied at inference time as well. A way to reduce such a problem is to enforce the constraints not only during training but also at inference time~\cite{SBR_collective}. However, notice that in this case the time complexity of inference is increased since the back-propagation needs to be used even in this case. On the other hand, methods that inject knowledge directly into the model structure are naturally capable of enforcing the knowledge at inference time.





Another possible advantage of model based methods is the possibility to learn a weight that codifies the importance of a logical rule directly from the data. This is no possible at all with methods based on regularization, since the logical formulas are directly codified inside the Loss function. As a consequence, if some rule weight is available, it must be added as a hyper-parameter.

Among the few methods that admit learnable weights associated with the logic formulas there is \kenn. This makes it suitable for scenarios where the given knowledge contains errors or when rules are softly satisfied in the real world but it is not known in advance the extent to which they are correct.
\kenn\ injects knowledge on top of the NN model through an additional layer which increases the satisfaction of the constraints in a fuzzy logic semantic.

Another approach
is provided by Li and Srikumar who recently proposed a method that codifies the logical constraints directly into the neural network model~\cite{augmenting}. However, they restrict the rules to implications with exactly one consequent and they do not provide the possibility to learn clause weights, which in their system are added as hyper-parameters. 

Going in the same direction, \cite{RNM} proposed Relational Neural Networks (RNM). RNM can be also inserted in the set of approaches that add the logic directly into the model and, as the best of our knowledge, it is the only method other than \kenn\ which is capable of integrating logical knowledge with a neural network while learning the clause weights. RNM integrates a neural network model with a FOL reasoner. This is done in two stages: in the first one, the NN is used to calculate initial predictions for the atomic formulas; in the second stage a graphical model is used to represent a probability distribution over the set of atomic formulas. To obtain the final predictions a Maximum a Posteriori (MAP) estimation is performed, finding the most probable assignment to the grounded atoms given the output of the NN and the set of constraints. At a high-level RNM approach is similar to \kenn, since in both cases a NN makes initial predictions and a post elaboration step is applied to such predictions to provide the final classification. However, RNM requires to solve an optimization problem at inference time and after each training step. This has the advantage of considering all the logical rules together at the same time at the expense of an increased computational effort. Contrary, in \kenn\ each rule is considered separately from the others, and the second stage is directly integrated inside the model as a differentiable function that can be trained end-to-end with the NN. However, 
with this strategy there could be some contradictory changes when combining multiple clauses with the same predicates. We will further analyze this aspect in \new{Section~\ref{sec:reasoning}, proposing a strategy to handle this limitation. Moreover, in Section~\ref{sec:collective}, we analyze this strategy empirically.}

%% file: kenn.tex
\section{Knowledge Enhanced Neural Networks}
We define the prior knowledge in terms of formulas of a function-free first order language $\L$. Its signature is defined with a set of domain constants $\C \triangleq \{a_1, a_2, ... a_m \}$
and a set of predicates $\P \triangleq \{P_1, P_2 ... P_q \}$.
In our setting, predicates can be unary or binary/
Binary predicates can express relations among pairs of objects in the domain, e.g. $Friends(a,b)$ states that person $a$ is a friend of $b$.
The prior knowledge is defined as a set of clauses: $\clauses \triangleq \{ c_1, c_2, ... c_r \} $. A clause is a disjunction of literals, each of which is a possibly negated atom:
	$
	c \triangleq \bigvee\limits_{i=1}^{k} l_i
	$, where $k$ is the number of literals in $c$ and $l_i$ is the $i^{th}$ literal. We assume that there are no repeated literals.
Since we are interested in representing only general knowledge, the literals do not contain any constant, only variables that are assumed to be universally quantified. If the predicate is binary, the two variables are $x$ and $y$, otherwise only $x$.
When an entire clause contains only variable $x$ (i.e., only unary predicates), we call it \emph{unary}. Similarly, if it contains both $x$ and $y$ we call it \emph{binary}\footnote{We restrict to the case where clauses contain at most two variables}.


As an example, the clause $\lnot Smoker(x) \lor Cancer(x)$ is unary and states that all smokers have also cancer (notice that the clauses are not assumed to be hard constraints). Instead, the clause
\begin{equation}
	\lnot Smoker(x) \lor \lnot Friends(x,y) \lor Smoker(y)
	\label{eq:clause}	
\end{equation}
is binary. It states that if a person $x$ is a smoker and he is a friend of another person $y$, then $y$ is also a smoker. We will use extensively this clause in the remaining of the paper, referring to it as $c_{SF}$.


We define the grounding of a unary clause $c$, denoted by $c[a]$, as the clause obtained by substituting the $x$ variable with constant $a$. Similarly, if $c$ is binary, its grounding $c[a,b]$ is obtained by substituting $x$ and $y$ with $a$ and $b$ respectively. For instance, 
the grounding $c_{SF}[a,b]$ of the clause defined in~ Eq.\ref{eq:clause} correspond to
$
\lnot Smoker(a) \lor \lnot Friends(a,b) \lor Smoker(b)
$.

\subsection{\kenn\ architecture}
Suppose we have a NN for a classification task which takes as input a matrix $\bx \in \Reals^{d \times n}$ containing $n$ features for $d$ samples, and returns an output $\by \in [0,1]^{d \times q}$ which contains the predictions for $q$ classes corresponding to the $q$ predicates.
A prior knowledge $\clauses$ is also provided. 
It can be used by \kenn\ to improve the predictions of the NN. 

\begin{figure}
	
	\centering
        \includegraphics[scale=0.16]{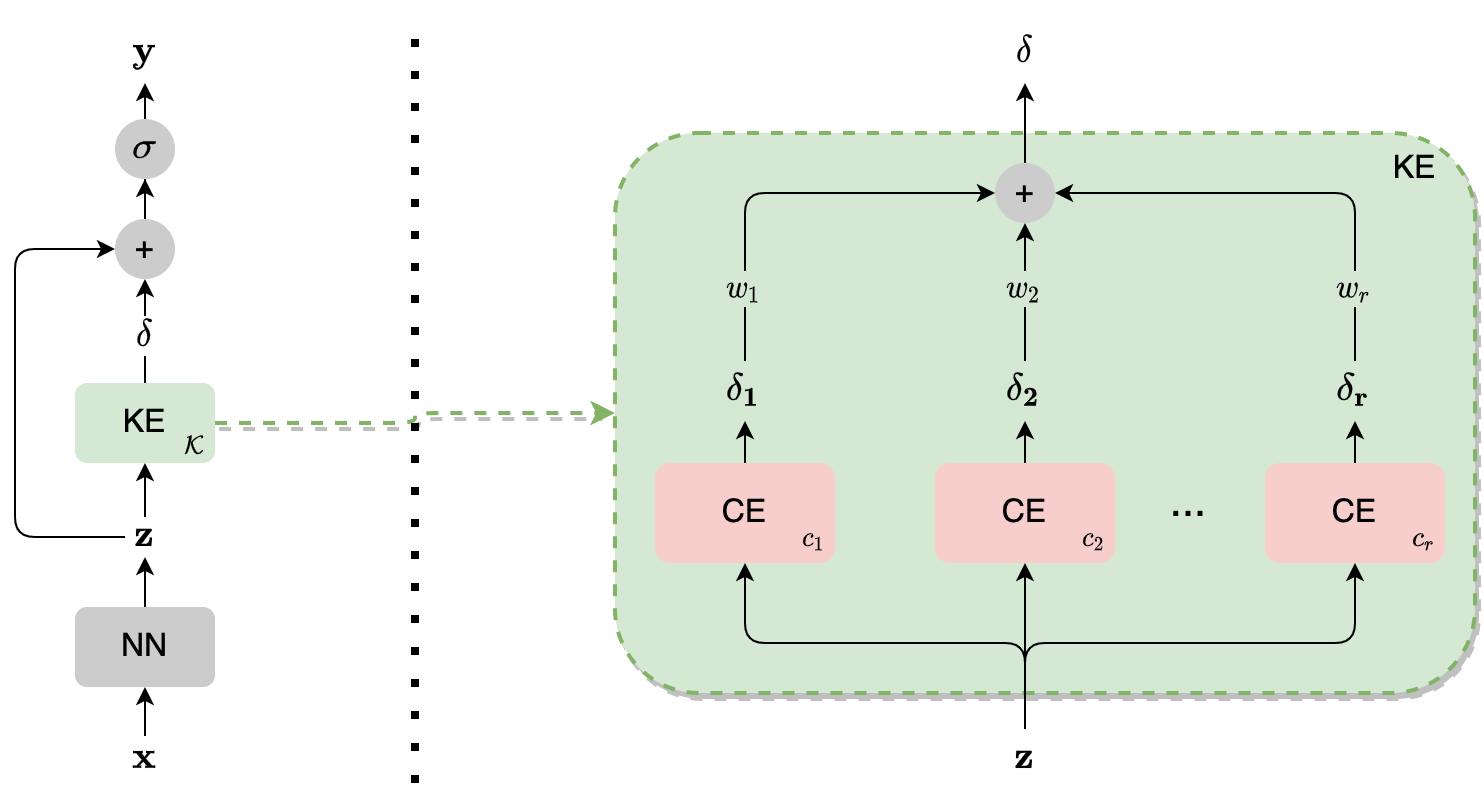}
	\caption{Model architecture. Left: \kenn\ model overview. 
	Right: Knowledge Enhancer.
	}
	\label{fig:KENN_overview}
\end{figure}

Figure~\ref{fig:KENN_overview}(left) shows a high-level overview of \kenn \ \new{where} a residual layer, called \emph{Knowledge Enhancer} (KE), is inserted between the NN and the final activation function. The role of KE is to revise the final predictions returned by the NN in order to increase the truth value of each clause $c \in \clauses$. It does so by calculating a residue $\bdelta$, a matrix that is added to the predictions of the NN. 

The KE works in the pre-activations space, i.e. on $\bz$, and the activation function ($\sigma$) is called later. In order for \kenn\ to work, the activation function must be monotonic and return values in the range $[0,1]$~\footnote{For more details on why the KE is applied on pre-activations, please refer to~\cite{kenn}}.
Since both NN and KE are differentiable, the entire architecture is differentiable end-to-end, making it possible to apply back-propagation algorithm on the whole model. Figure~\ref{fig:KENN_overview}(right) shows the architecture of KE, which calculates the residual matrix $\bdelta$.



More in details, for each clause $c\in\clauses$, the KE contains a submodule, the \emph{Clause Enhancer} (CE), which proposes the changes $\bdelta_c$ to be applied on the NN's preactivations in order to increase the satisfaction of $c$. Indeed, the CE computes a soft differentiable approximation of a function called \emph{t-conorm boost function} (TBF). Intuitively, a TBF is a function \mbox{$\phi: \Reals^k \to \Reals_+^k$} that proposes the changes to be applied on the pre-activations $\bz$ of $k$ truth values, such that $\bot(\sigma(\bz + \phi(\bz))) \geq \bot(\sigma(\bz))$, where $\bot: [0,1]^k \to [0,1]$ is a t-conorm function,
used in fuzzy logic to represent the semantics of the disjunction operator~\footnote{In \cite{kenn}, function $\phi$ is called $\delta$. Here we changed the name to avoid confusions with its output which is also referred as $\delta$}. In~\cite{kenn} it has been defined the function 
\begin{equation}
	\phi(\bz)_i = 
	\left\{
	  \begin{array}{ll}
	1 \ \ \ & \mbox{if $i = \argmax_{j=1}^n z_j$} \\
	0 & \mbox{otherwise}
	\end{array}
	\right.
	\label{eq:phi}
\end{equation}
and proved that such a function is the optimal TBF for
the G\"{o}del t-conorm. 
\kenn\ employs the softmax function
as a continuous and differentiable approximation of $\phi$.
 
The $\bdelta_c$ matrices are combined linearly inside the KE to obtain the final change $\bdelta$ to be applied on the NN's predictions, and finally the $\bdelta$ is summed to the initial pre-activations $\bz$ and passed to the activation function:

\begin{equation}
	\begin{split}
	y_{P(a)} &= \sigma \bigl(z_{P(a)} + \sum_{\substack{c \in \clauses\\ P(x) \in c}} w_c \cdot \delta_{c[a],P(a)}\bigr)
	\end{split}
	\label{eq:final_intepretation_unary}
\end{equation}

where $w_c$ is the \emph{clause weight}, $P(a)$ a grounded atom, $y_{P(a)}$ its final prediction, and $z_{P(a)}$ the NN's pre-activations. Finally, $\delta_{c[a],P(a)}$ is the change applied to $P(a)$ based on the grounded clause $c[a]$:
\begin{equation}
    \label{eq:deltac}
  \delta_{c[a],P(a)}=\begin{cases}
    \phi(\bz_c)_{P(a)} & \mbox{if $P(a)\in c[a]$} \\
    -\phi(\bz_c)_{\lnot P(a)} & \mbox{if $\neg {P(a)}\in c[a]$} 
  \end{cases}
\end{equation}
where $\bz_c$ are the pre-activations of literals of $c$.

\new{
Note that applying a linear combination of the $\bdelta_c$ matrices can be done under the assumption of independence between the clauses. 
When multiple clauses share common predicates
the changes proposed by \kenn\ could only partially improve the satisfaction of the knowledge. We will further analyze this problem in Section~\ref{sec:reasoning}}.


%% file: relational.tex
\subsection{Extending \kenn\ for relational domains}
\label{sec:kenn_extension}



\begin{figure}
	\centering
        \includegraphics[scale=0.13]{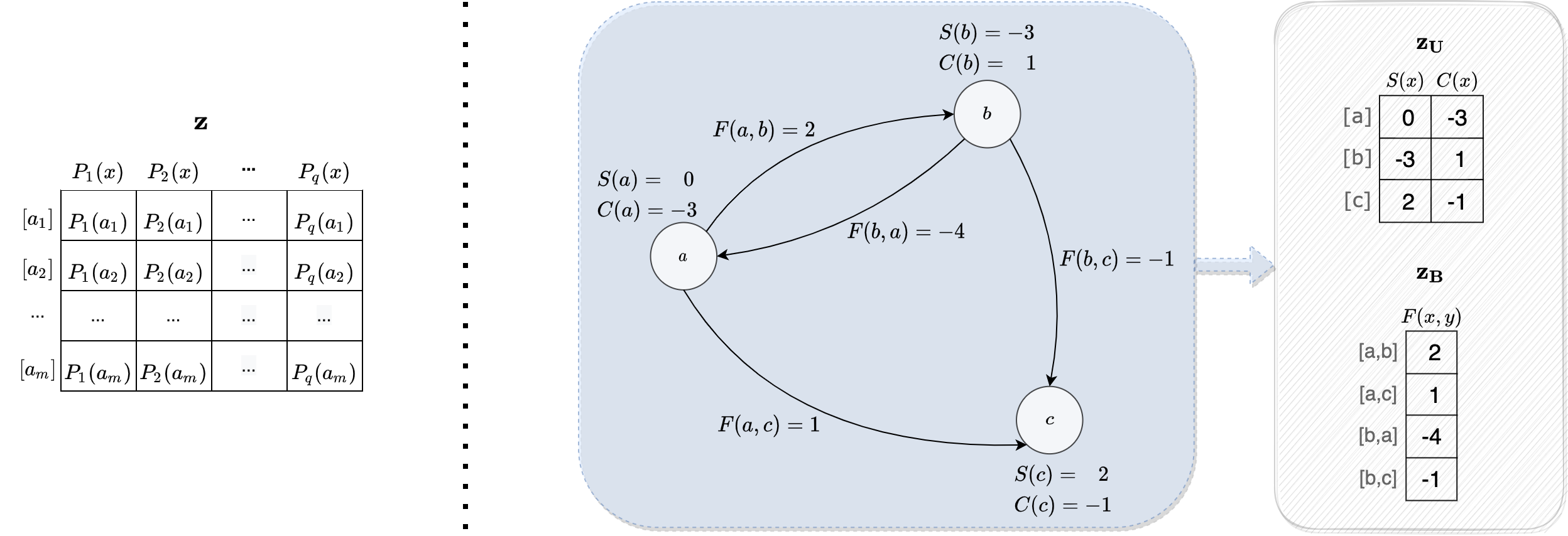}
	\caption{The representation of NN's final pre-activations. Left: unary case. Right: representation of relational data. Pre-activations are represented as integers instead of reals to simplify the figure.}
	\label{fig:z}
\end{figure}

In the architecture defined so far, the groundings involve a single object and $\bz$ is defined as a matrix, where columns represent predicates and rows constants. Fig.~\ref{fig:z}(left) introduces 
the representation of $\bz$: it is defined as a matrix such that the element $z_{ij}$ contains the pre-activation of $P_j(a_i)$, with $P_j$ the $j^{th}$ predicate and $a_i$ the $i^{th}$ constant.

Note that this kind of representation is common when working with neural networks since the columns (predicates) correspond to the labels and the rows (groundings) to the samples. An important aspect of this representation lies in the fact that each grounded atom can be found in the matrix exactly one time. This allows to parallelize computations of Equation~\ref{eq:final_intepretation_unary} since a grounded clause involves only atoms in the same row, and each row can be managed in parallel inside a GPU. This can be done only if the same atom does not appear in multiple rows, since the changes are applied independently to each row and are not aggregated together. This property always holds with unary clauses.



To represent relational data, we extend \kenn\ with an extra matrix $\bz_B$, which contains the binary predicates' pre-activations. For uniformity of notation we use $\bz_U$ to denote the unary matrix $\bz$ of the not relational \kenn.
Matrix $\bz_B$ contains one row for every pair of objects we are interested in and a column for each binary predicate. 
Fig.~\ref{fig:z}(right) shows this representation using the classical Smoker-Friends-Cancer example, where the domain is composed of three constants (persons) $\C = \{a, b, c\}$, the unary predicates are $S$ and $C$ (for $Smoker$ and $Cancer$), and a binary predicate $F$ (for $Friends$). The blue box shows the graph representation with nodes and edges labelled with pre-activation of unary and binary predicates respectively. The grey box shows the corresponding matrix representation used by  \kenn. Notice that it is not required that the entire graph is computed by the NN. For instance, in the experiments on Citeseer, the $Cite$ predicate is provided directly as a feature (see Section~\ref{sec:collective}).


The architecture of \kenn\ for relational domains is
very similar to the architecture of traditional \kenn\ of Fig.~\ref{fig:KENN_overview}, with KE substituted by a \emph{Relational KE} (RKE). From a high level perspective, the RKE differs from the traditional KE on the amount of inputs and outputs. As seen before, in the relational case the pre-activations are divided in two different matrices ($\bz_U$ and $\bz_B$) and, as a consequence, also the $\bdelta$ matrix and predictions $\by$ are now splitted in unary and binary matrices ($\bdelta_U$ and $\bdelta_B$ for the residues, $\by_U$ and $\by_B$ for the final predictions).

The RKE has the same role as the KE in the unary case. However, it is capable to consider also binary predicates. When binary knowledge is available, additional steps are required since the independence between object can not be assumed anymore.

Let $\clauses_U$ be the set of unary clauses and $\clauses_B$ the set of binary clauses. The prior knowledge is now defined as $\clauses = \clauses_U \cup \clauses_B$.
The idea is to apply the KE to these two sets separately. Equation~\ref{eq:final_intepretation_unary} can be decomposed using the new defined partition of the knowledge:
$
	y_A = \sigma \left( z_A + \sum_{c \in \clauses_U[\C]} w_c \cdot \delta_{c,A} + \sum_{c \in \clauses_B[\C]} w_c \cdot \delta_{c,A} \right)
$, 
where $A$ is a grounded atom (i.e. $P(a)$ or $P(a,b)$, depending on the arity of $P$).
We define $\bdelta_{\clauses_U}$ as the changes deriving from unary clauses:
\begin{equation}
    \delta_{\clauses_U,P(a)} = \sum_{\substack{c \in \clauses_U\\ P(x) \in c}} w_c \cdot \delta_{c[a],P(a)}
	\label{eq:ku_pa}
\end{equation}
Similarly, $\bdelta_{\clauses_B}$ are the changes calculated from $\clauses_B$.

Notice that the approach defined so far can be directly applied to the unary knowledge $\clauses_U$ to calculate $\bdelta_U$ since traditional KE can manage unary knowledge. Indeed, internally the RKE contains a standard KE which manages the unary clauses. 

We need to define a strategy to deal with binary clauses. Indeed, when a clause $c$ contain two variables, 
a grounding of a unary predicate may occur in multiple groundings of $c$.
For instance, consider the clause of Eq.~\ref{eq:clause}.
The two groundings $c_{SF}[a,b]$ and $c_{SF}[b,c]$ share a common grounded atom: $Smoker(b)$. 
For this reason, when dealing with the predictions of a unary predicate in a relational domain, we need to account for such repetitions:
\begin{equation}
    \delta_{\clauses_B,P(a)} = \sum_{b \neq a} \Big( \sum_{\substack{c \in \clauses_B\\ P(x) \in c}} w_c \cdot \delta_{c[a,b],P(a)} +  \sum_{\substack{c \in \clauses_B\\ P(y) \in c}} w_c \cdot \delta_{c[b,a],P(a)} \Big)
	\label{eq:kb_pa}
\end{equation}

Putting all together, the predictions $y_{P(a)}$ for a grounded unary predicate $P$ are defined as:
\begin{equation}
	y_{P(a)} = \sigma(z_{P(a)} + \delta_{\clauses_U,P(a)} + \delta_{\clauses_B,P(a)} )
	\label{eq:y_pa}
\end{equation}

The predictions for a binary predicate $R$ are easier to be computed since they can be found only in binary clauses and any possible grounding of $R$ can be found in only one corresponding grounding of each clause
$y_{R(a,b)} = \sigma(z_{R(a,b)} + \delta_{\clauses_B,R(a,b)})$, 
with
\begin{equation}
	\delta_{\clauses_B,R(a,b)} = \sum_{\substack{c \in \clauses_B\\ R(x,y) \in c}} w_c \cdot \delta_{c[a,b],R(a,b)}
	\label{eq:kb_rab}	
\end{equation}

 
\subsection{\new{Time complexity}}
\new{Here we analyze the time complexity of an RKE layer with respect to domain size $m$, number of predicates $|\P|$, and number of rules $|\clauses|$. We also assume the maximum number $L$ of literals in a clause to be a small constant.
	
Let us first analyze the time complexity for calculating the $\delta_{c[a]}$ used in Equations~\ref{eq:ku_pa}, and \ref{eq:kb_pa}. Each $\delta_{c[a],P(a)}$ can be calculated in time $O(1)$ (see Equation~\ref{eq:deltac}). 
Computing $\delta_{c[a]}$ also requires constant time. The sum of Equation~\ref{eq:ku_pa} require time $O(|\clauses|)$, which is the time necessary to compute $\delta_{\clauses_U,P(a)}$. Note that neural networks are usually run on GPUs, where the computations can be parallelized. Assuming enough parallel processes ($|\clauses|$ in this case), a sum can be performed in a time logarithmic with respect to the number of addends, and complexity for $\delta_{\clauses_U,P(a)}$ becomes $O(log(|\clauses|))$. Finally, Equation~\ref{eq:ku_pa} needs to be calculated for all the grounded unary predicates $P(a)$, for a total time of $O(m \cdot |\P| \cdot |\clauses|)$ in a single process, and $O(log(|\clauses|))$ with multiple parallel processes (each of the grounded atom can be considered independently from the others). With a similar reasoning, we found the time complexity of Equations~\ref{eq:kb_pa} and \ref{eq:kb_rab} to be $O(m^2 \cdot |\P| \cdot |\clauses|)$. Note that with enough parallel processes we can compute all the deltas in $O(log(m) + log(|\clauses|))$.

}

\subsection{Treatment of dependencies among the rules}
\label{sec:reasoning}
\new{In the previous section we showed the efficacy of the method in terms of execution time, which can be achieved thanks to the assumption of independence. However, when this assumption is violated, \kenn\ does not provide any guarantees on the satisfaction of the knowledge.}
\new{As an example,} suppose that we have two grounded clauses $c_1: \lnot A \lor B$ and $c_2: \lnot B \lor C$ with their respective clause enhancers CE$_1$ and CE$_2$, where $A$, $B$ and $C$ are grounded unary or binary predicates. The atom $B$ appears in both clauses with opposite signs.
Since the CE increase the highest literal value (see Eq.~\ref{eq:phi}), if $A < B$~\footnote{With an abuse of notation, we use the atoms symbols to refer also to their truth value} and $C < \lnot B$, then CE$_1$ increases $B$ and CE$_2$ decreases it. As a consequence, the satisfaction of only one between $c_1$ and $c_2$ is increased. Furthermore, the satisfaction of the entailed clause $\lnot A \lor C$ is also not improved. 


For any grounded atom $G$, lets define $G^{(0)}$ as its initial prediction and $G^{(i)}$ as the prediction of the $i^{th}$ KE layer. Moreover, suppose that all KEs share the same clause weights $w_1$ and $w_2$ (for $c_1$ and $c_2$ respectively). From Equations~\ref{eq:final_intepretation_unary} and \ref{eq:deltac} we can derive $B^{(1)} = B^{(0)} + w_1 - w_2$, and $\lnot B^{(1)} = \lnot B^{(0)} + w_2 - w_1$. If $w_1 \geq w_2$, then 
$A^{(1)} = A^{(0)} < B^{(0)} \leq B^{(1)}$.
As a consequence, the first rule will increase again $B$ even at the next KE layer. On the other hand, the value of $\lnot B$ is reduced, which means that there is an increased chance for $C^{(1)} > \lnot B^{(1)}$, which would solve the problem since CE$_2$ would increase $C$ instead of $\lnot B$. Notice that, since the weights are the same at each level, it is always true that $\lnot B^{(i+1)} \leq \lnot B^{(i)}$, meaning that with enough KE layers both clauses' satisfaction will be increased (and as a consequence, also their entailments).

\new{The problem analyzed in this section } becomes even more relevant in relational domains since in these contexts an atom can be shared not only by multiple clauses but also by different groundings of the same clause (for instance, in $c_{SF}[a,b]$ and $c_{SF}[b,c]$). For this reason, in these contexts stacking multiple RKEs is recommended (more details in Section~\ref{sec:results}).

%% file: collective.tex
\section{Evaluation of the model}\label{sec:collective}

In this section, the relational extension of \kenn\ is tested on the task of Collective Classification: given a graph,
we are interested in finding a classification for its nodes using both features of the nodes (the objects) and the information coming from the edges of the graph (relations between objects)~\cite{collective_classification}.

In Collective Classification, there are two different learning tasks: inductive and transductive learning. In inductive learning, there are two separate graphs, one for training and the other for testing. On the contrary, in transductive learning, there is only one graph that contains nodes both for training and testing. In other words, in inductive learning, there are no edges between nodes for training and testing, while in transductive learning there are. 
The tests have been performed on both tasks \new{to analyze the behavior of \kenn\ in the contexts of relational domains. In particular, we tested \kenn\ with a varying number of KEs layers to validate the proposal of Section~\ref{sec:reasoning}.\footnote{Source code of the experiments are available on \href{https://github.com/rmazzier/KENN-Citeseer-Experiments}{https://github.com/rmazzier/KENN-Citeseer-Experiments}
}. }

\subsection{Experimental setup}
\label{sec:exp_setup}
We followed the evaluation methodology of~\cite{RNM}, where the experiments have been carried out on Citeseer dataset~\cite{citeseer} using SBR and RNM. 
The Citeseer dataset used in the evaluation is a citation network: the graph's nodes represent documents and the edges represent citations. The nodes' features are bag-of-words vectors, where an entry is zero if the corresponding word of the dictionary is absent in the document, and one if it is present. The classes to be predicted represent possible topics for a document.
The dataset contains 3312 nodes that must be classified in 6 different classes: AG, AI, DB, IR, ML, and HCI.
The classification is obtained from the 3703 features of the nodes, with the addition of the information coming from the citations (4732 edges).


\new{
We use the same NN and knowledge as in~\cite{RNM}, allowing for the comparison with SBR and RNM.}
The NN is a dense network with 3 hidden layers, each with 50 hidden nodes and ReLU activation function. 
\new{The knowledge consists of six rules obtained by substituting the topic $T$ in $
\lnot T(x) \lor \lnot Cite(x,y) \lor T(y)
$ with all the classes, and they codify the idea that papers cite works that are related to them. 
}

The training set dimension is changed multiple times to evaluate the efficacy of the three methods on the varying of training data. More precisely, tests have been conducted by selecting 10\%, 25\%, 50\%, 75\%, and 90\% of nodes for training. For each of these values, the training and evaluation were performed 100 times, each with a different split of the dataset. At each run the training set is created by selecting random nodes of the graph, with the constraints that the dataset must be balanced.

\subsection{Results}
\label{sec:results}
\begin{figure}
    \centering
        \includegraphics[scale=0.35]{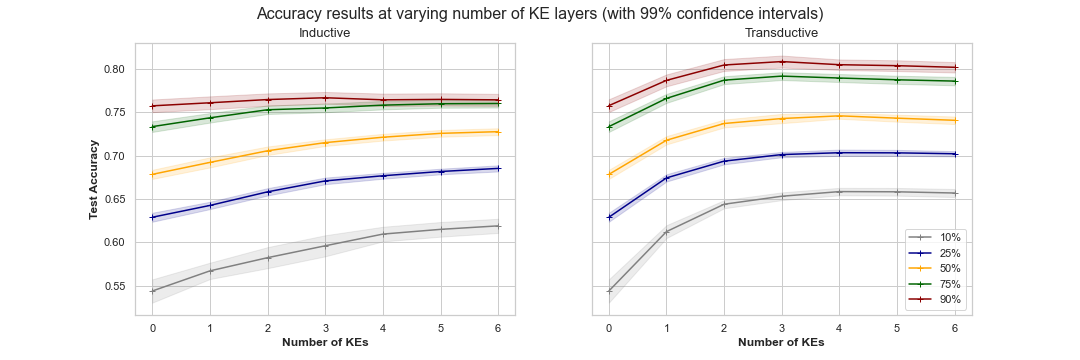}
    \caption{\new{Accuracies of \kenn\ at the varying of KEs layers.
    }}
    \label{fig:comparison_n_layers}
\end{figure}

\new{Figure~\ref{fig:comparison_n_layers} shows the test accuracies obtained by \kenn\ while increasing the number of KEs layers, starting from $0$ (corresponding to the NN accuracy) up to 6. Note that, for each line in the figure, there is a surrounding border corresponding to a 99\% confidence interval.
To calculate the intervals, we assumed the distribution of improvements obtained by the injection of the logical rules to be a normal distribution (see figures in Appendix B and C).
We also computed the p-values for each setting, assuming as null Hypothesis 
that the distribution of accuracies of the NN is the same as \kenn. Since the number of runs, as well as the obtained improvements, are quite high, the resulting p-values are very small. For this reason, we can safely reject the null Hypothesis, and we are very confident that the improvements given by \kenn\ do not depend on the random initialization of the models' parameters or the specific choices of the splits of the dataset.}

\new{More in detail, we found p-values in the range from 8.2e-42 to 1.6e-09 in the inductive case, and from 53e-72 to 2.1e-23 for the transductive one. The only exception is with 90\% of the samples in the inductive scenario where the p-value is 0.35. This is because the improvements over the NN are very small. Indeed, in both learning paradigms, the effect of the knowledge is reduced when the amount of available data is larger. This behavior is consistent with the simple intuition that, when the training data is scarce, the usage of knowledge should bring higher benefits.
} 

\new{A more important result coming from these experiments is the fact that in all cases adding a new KE layer does not reduce the test accuracy. On the contrary, most of the time the metric is increased until a certain number of layers is reached, and after that, the accuracy stabilizes. This behavior is in line with the discussion of Section~\ref{sec:reasoning} and confirms the efficacy of the proposed strategy to deal with the violation of independece assumption. }


\new{
Finally, Figure~\ref{fig:comparison_n_layers} provide also a measure of the amount of information carried out by the knowledge. For instance, consider blue and yellow lines, corresponding to a training set with 25\% and 50\% of the samples, respectively. In the inductive scenario, the accuracy obtained with 25\% with the addition of the knowledge is almost the same as the standard NN with 50\% of the data (even higher in the transductive scenario). In this case, adding the knowledge has the same effect of doubling up the training data! Indeed, one of the main motivations behind Neural-Symbolic Integration consists in reducing the required amount of training samples since collecting labeled data is costly in practice.
}


\subsection{Comparison with other NeSy frameworks}
\label{sec:comparison}
\new{Table~\ref{tab:comparison} shows a comparison of \kenn\ with SBR and RNM. We used the results of \kenn\ with 3 KEs since more layers do not provide a significant advantage (see Section~\ref{sec:results}).
}

\begin{table}[H]
    \caption{Improvements in terms of accuracy on inductive and transductive learning. 
    }
    \centering
    \begin{tabular}{c|lll|lll}
        \multicolumn{4}{c}{\ \ \ \ \ \ \ Inductive} & \multicolumn{3}{c}{Transductive} \\
        \rule{0pt}{2ex}\hspace{-0.12cm}
        \% Tr & SBR & RNM & \kenn & SBR & RNM & \kenn \\
        \hline
        \rule{0pt}{2ex}\hspace{-0.12cm}
        10 & 0.005 & 0.040 & {\bfseries 0.052} & 0.063 & 0.068 & {\bfseries 0.110}\\
        \rule{0pt}{1.5ex}\hspace{-0.12cm}
        25 &  0.008 & 0.035 & {\bfseries 0.044} & 0.062 & 0.068 & {\bfseries 0.074}\\
        \rule{0pt}{1.5ex}\hspace{-0.12cm}
        50 & 0.005 & 0.019 & {\bfseries 0.036} & 0.052 & 0.058 & {\bfseries 0.064} \\
        \rule{0pt}{1.5ex}\hspace{-0.12cm}
        75 & 0.002 & 0.009 & {\bfseries 0.021} & 0.056 & {\bfseries 0.058} & 0.057 \\
        \rule{0pt}{1.5ex}\hspace{-0.12cm}
        90 & 0.003 & {\bfseries 0.009} & 0.001 & {\bfseries 0.054} & {\bfseries 0.054} & 0.043\\
        \hline
        \hline
    \end{tabular}
    \label{tab:comparison}
\end{table}

\new{As we can see from the table, in the inductive case SBR produces much lower improvements compared to the other two methods.} Note that these results are in line with previous results obtained on VRD dataset, where another regularization approach (LTN) was compared with KENN~\cite{kenn}. Indeed, the results obtained in both VRD and Citeseer suggest better performances of model-based approaches as compared to the ones based on regularization. 
Note that methods based on regularization of the loss do not impose the knowledge at inference time. In the transductive scenario, the situation is different and SBR behaves similarly to the other two. Indeed, in this case, citations between training and test nodes are available and there is no distinction between training and inference.

\new{
Finally, the results suggest that \kenn\ is particularly useful when the training data available is scarce. On the contrary, when data is abundant, our results tend to degrade faster than RNM and SBR.
However, the greatest advantage of \kenn\ over other architectures is its scalability. This is confirmed by the comparison of the execution times of the three methods: we found \kenn\ to be very fast as compared to the other two methods with an average of 7.96s required for a single run, as compared to the NN which requires 2.46s (on average of 1.83 seconds for each KE layer). A run of SBR cost 87.36s (almost 11 times slower than \kenn), while RNM required 215.69s per run (27 times slower)~\footnote{All the experiments have been run on the same architecture, an NVIDIA Tesla v100}.}

\section{Conclusions}
\kenn\ is a NeSy architecture that injects prior logical knowledge inside a neural network by stacking a residual layer on its top.
In~\cite{kenn}, it proved to be able to effectively inject knowledge in the context of multi-label classification tasks. In this work, we extended \kenn\ for relational domains, where the presence of both unary and binary predicates doesn't allow for the usage of the simple tabular representation of the data used in the previous version of the framework. \new{Moreover, we propose a strategy to deal with the violation of the independence assumption made by \kenn.} The experiments on Citeseer show \new{the effectiveness of this strategy, obtaining} statistically relevant improvements over the NN performances, meaning that \kenn\ can successfully inject knowledge even in the presence of relational data. Finally, \kenn\ provided quality results also in comparison with other two NeSy frameworks. In particular, the large difference in performances between \kenn/RNM and SBR provides additional evidence in support of model-based approaches in comparison to regularization ones\new{, with \kenn\ the best option in terms of scalability.}

%% file: appendix.tex
\section*{Appendix}
\appendix

\section{Relational \kenn\ architecture}
\label{ap:rel_kenn}
\begin{figure}[H]
	\centering
\includegraphics[scale=0.1]{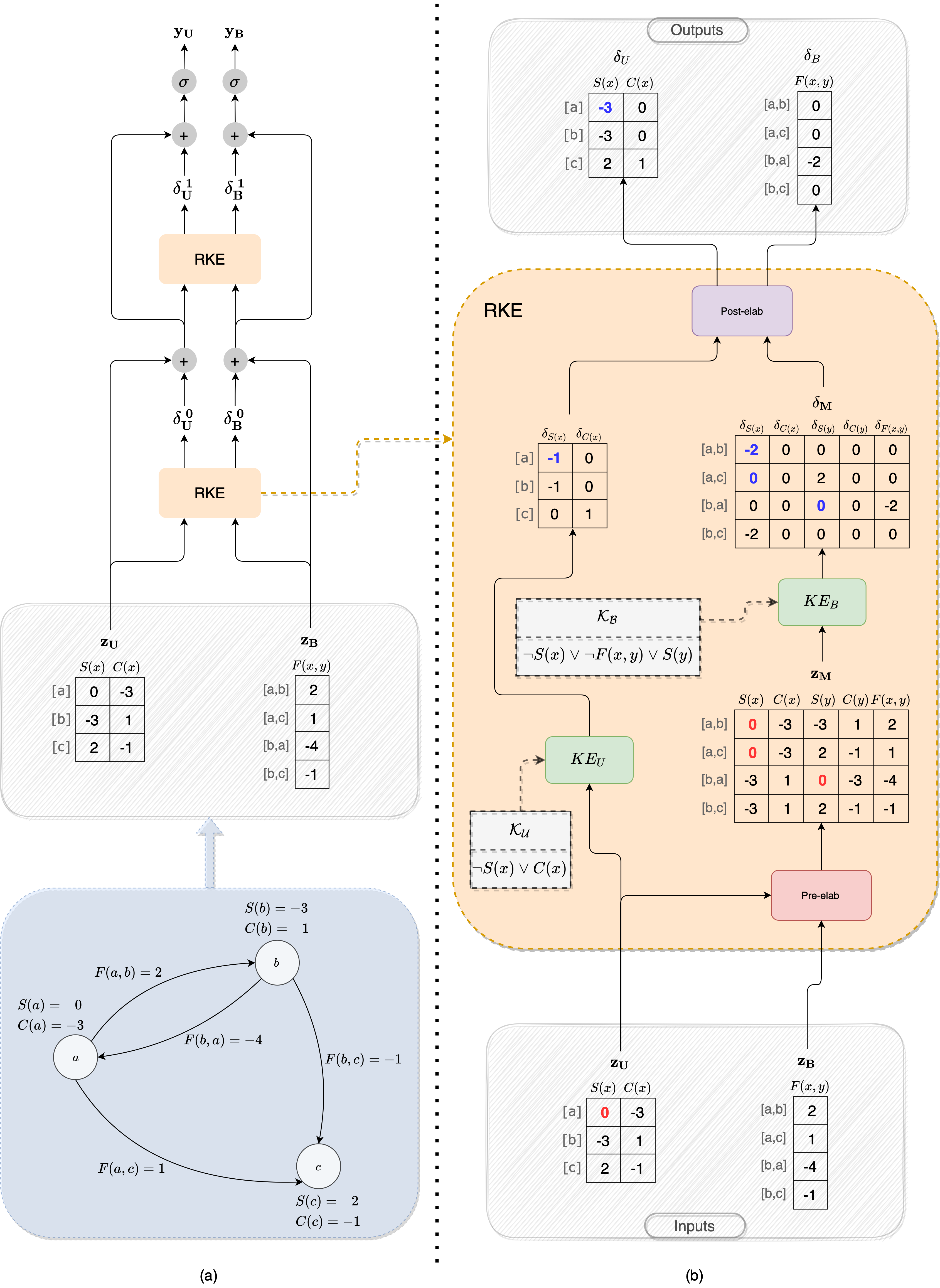}

    \caption{
        \kenn\ for relational domains: (a) the architecture of \kenn. A graph (blue box) is represented in terms of the two matrices $\bz_U$ and $\bz_B$ and given as input to the Relational KE (RKE). Multiple RKEs are stacked together and the activation function is called; (b) the architecture of the RKE module: the unary knowledge is enforced directly by the KE$_U$; the binary knowldge is enforced by the KE$_B$ on matrix $\bz_M$, which is created by joining $\bz_U$ with $\bz_B$ on the pre-elab step. $\bz_M$ contains multiple instances of the same atoms, for instance $S[a]$ (red cells). As a consequence, multiple residues are returned for a single atom, and such values are summed in the post-elab (blue cells). Pre and post elaboration steps are efficiently implemented using TensorFlow \emph{gather} and \emph{scatter\_nd} functions.
        }
    \label{fig:relational_kenn}
\end{figure}

\section{Results distribution - Inductive Learning}
\label{ap:inductive}
\begin{figure}[H]
    \makebox[\textwidth][c]{
        \includegraphics[scale=0.55]{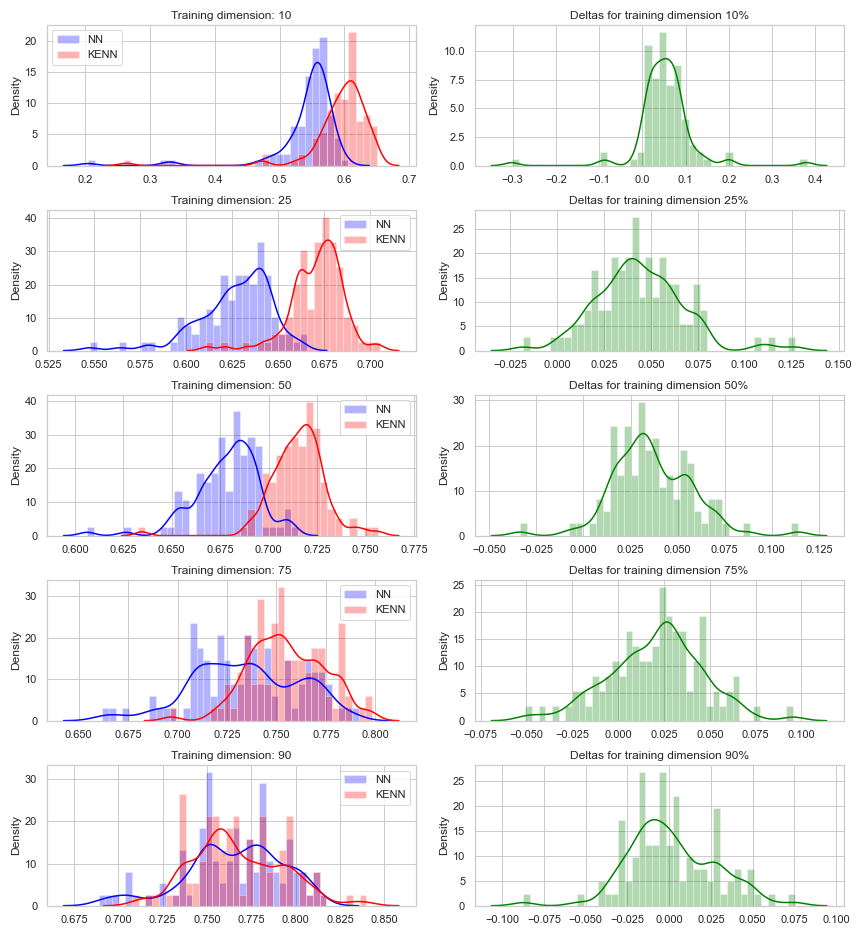}
    }
    \caption{Left: distributions of accuracies achieved by the NN and KENN (3 KE layers) on 100 runs of Inductive Learning; Right: distributions of the improvements in accuracy obtained by the injection of the logical rules.}
    \label{fig:inductive_histograms}
\end{figure}

\section{Results distribution - Transductive Learning}
\label{ap:transductive}

\begin{figure}[H]
    \makebox[\textwidth][c]{
        \includegraphics[scale=0.55]{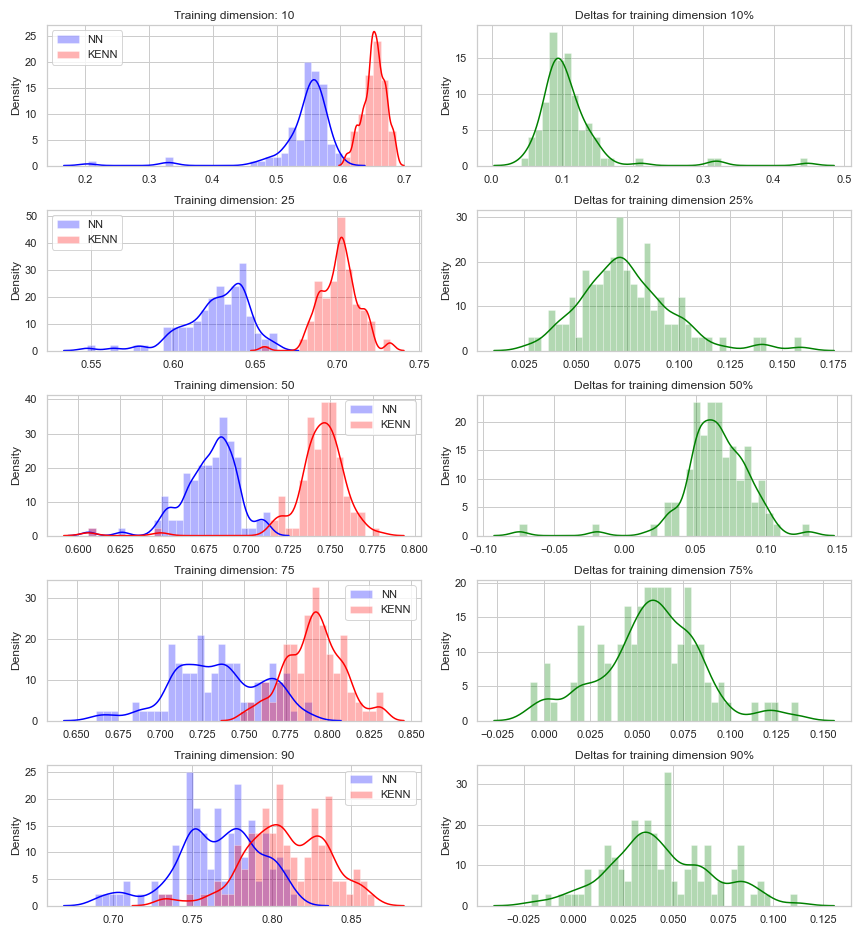}
    }
    \caption{Left: distributions of accuracies achieved by the NN and KENN (3 KE layers) on 100 runs of Transductive Learning; Right: distributions of the improvements in accuracy obtained by the injection of the logical rules.}
    \label{fig:inductive_histograms}
\end{figure}

\newpage
\section{Comparison with SBR and RNM}
\subsection*{Test accuracy}
\begin{figure}[H]
    \centering
        \includegraphics[scale=0.4]{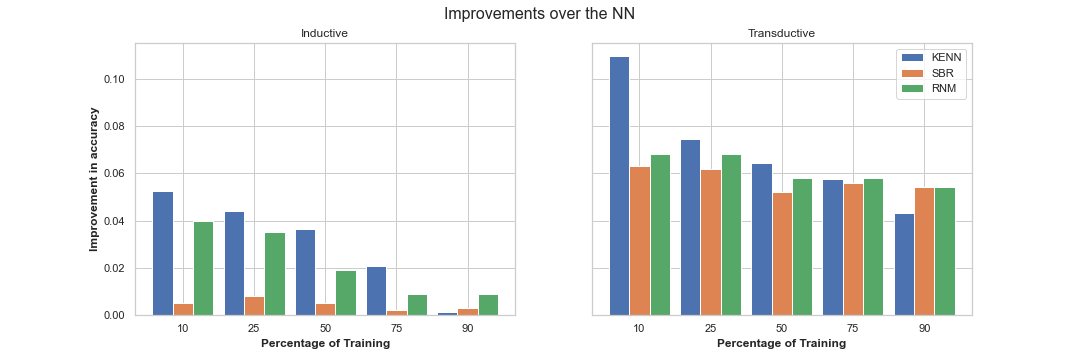}
    \caption{Comparison between KENN (3 KE layers), SBR and RNM in terms of accuracy improvements over the NN.}
    \label{fig:comparison}
\end{figure}

\subsection*{Execution time}
\begin{figure}[H]
    \centering
        \includegraphics[scale=0.4]{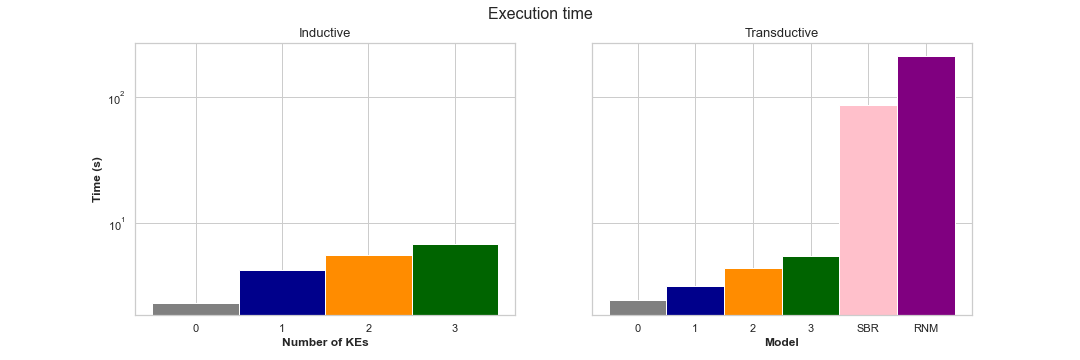}
    \caption{Execution time in logarithmic scale of the different methods. A bar labelled with number $i$ corresponds to \kenn\ with $i$ KEs layers ($0$ represents the NN without logic).}
    \label{fig:comparison_time_all}
\end{figure}

%% file: KENN_relational.bbl
\begin{thebibliography}{34}
\providecommand{\natexlab}[1]{#1}
\providecommand{\url}[1]{\texttt{#1}}
\expandafter\ifx\csname urlstyle\endcsname\relax
  \providecommand{\doi}[1]{doi: #1}\else
  \providecommand{\doi}{doi: \begingroup \urlstyle{rm}\Url}\fi

\bibitem[Bach et~al.(2017)Bach, Broecheler, Huang, and Getoor]{PSL}
S.~H. Bach, M.~Broecheler, B.~Huang, and L.~Getoor.
\newblock Hinge-loss {M}arkov random fields and probabilistic soft logic.
\newblock \emph{Journal of Machine Learning Research}, 18\penalty0
  (109):\penalty0 1--67, 2017.

\bibitem[Bahdanau et~al.(2014)Bahdanau, Cho, and Bengio]{machine_translation}
D.~Bahdanau, K.~Cho, and Y.~Bengio.
\newblock Neural machine translation by jointly learning to align and
  translate.
\newblock \emph{arXiv preprint arXiv:1409.0473}, 2014.

\bibitem[Besold et~al.(2017)Besold, d'Avila Garcez, Bader, Bowman, Domingos,
  Hitzler, K{\"{u}}hnberger, Lamb, Lowd, Lima, de~Penning, Pinkas, Poon, and
  Zaverucha]{neural_symbolic}
T.~R. Besold, A.~S. d'Avila Garcez, S.~Bader, H.~Bowman, P.~M. Domingos,
  P.~Hitzler, K.~K{\"{u}}hnberger, L.~C. Lamb, D.~Lowd, P.~M.~V. Lima,
  L.~de~Penning, G.~Pinkas, H.~Poon, and G.~Zaverucha.
\newblock Neural-symbolic learning and reasoning: {A} survey and
  interpretation.
\newblock \emph{CoRR}, abs/1711.03902, 2017.
\newblock URL \url{http://arxiv.org/abs/1711.03902}.

\bibitem[Campero et~al.(2018)Campero, Pareja, Klinger, Tenenbaum, and
  Riedel]{rule_induction}
A.~Campero, A.~Pareja, T.~Klinger, J.~Tenenbaum, and S.~Riedel.
\newblock Logical rule induction and theory learning using neural theorem
  proving.
\newblock \emph{arXiv preprint arXiv:1809.02193}, 2018.

\bibitem[Cohen(2016)]{tensorlog}
W.~W. Cohen.
\newblock Tensorlog: A differentiable deductive database.
\newblock \emph{arXiv preprint arXiv:1605.06523}, 2016.

\bibitem[Daniele and Serafini(2019)]{kenn}
A.~Daniele and L.~Serafini.
\newblock Knowledge enhanced neural networks.
\newblock In A.~C. Nayak and A.~Sharma, editors, \emph{PRICAI 2019: Trends in
  Artificial Intelligence}, pages 542--554, Cham, 2019. Springer International
  Publishing.
\newblock ISBN 978-3-030-29908-8.

\bibitem[De~Raedt et~al.(2007)De~Raedt, Kimmig, and Toivonen]{problog}
L.~De~Raedt, A.~Kimmig, and H.~Toivonen.
\newblock Problog: A probabilistic prolog and its application in link
  discovery.
\newblock In \emph{IJCAI}, volume~7, pages 2462--2467. Hyderabad, 2007.

\bibitem[Detassis et~al.(2021)Detassis, Lombardi, and Milano]{old_dog}
F.~Detassis, M.~Lombardi, and M.~Milano.
\newblock Teaching the old dog new tricks: Supervised learning with
  constraints.
\newblock In \emph{Thirty-Fifth {AAAI} Conference on Artificial Intelligence,
  {AAAI} 2021, Thirty-Third Conference on Innovative Applications of Artificial
  Intelligence, {IAAI} 2021, The Eleventh Symposium on Educational Advances in
  Artificial Intelligence, {EAAI} 2021, Virtual Event, February 2-9, 2021},
  pages 3742--3749. {AAAI} Press, 2021.
\newblock URL \url{https://ojs.aaai.org/index.php/AAAI/article/view/16491}.

\bibitem[Diligenti et~al.(2017)Diligenti, Gori, and Sacc{\`{a}}]{SBR}
M.~Diligenti, M.~Gori, and C.~Sacc{\`{a}}.
\newblock Semantic-based regularization for learning and inference.
\newblock \emph{Artif. Intell.}, 244:\penalty0 143--165, 2017.

\bibitem[Donadello(2018)]{IvanThesis}
I.~Donadello.
\newblock \emph{Semantic Image Interpretation - Integration of Numerical Data
  and Logical Knowledge for Cognitive Vision}.
\newblock PhD thesis, Trento Univ., Italy, 2018.

\bibitem[Dong et~al.(2019)Dong, Mao, Lin, Wang, Li, and Zhou]{NLM}
H.~Dong, J.~Mao, T.~Lin, C.~Wang, L.~Li, and D.~Zhou.
\newblock Neural logic machines.
\newblock \emph{arXiv preprint arXiv:1904.11694}, 2019.

\bibitem[Evans and Grefenstette(2018)]{dilp}
R.~Evans and E.~Grefenstette.
\newblock Learning explanatory rules from noisy data.
\newblock \emph{Journal of Artificial Intelligence Research}, 61:\penalty0
  1--64, 2018.

\bibitem[Fischer et~al.(2019)Fischer, Balunovic, Drachsler-Cohen, Gehr, Zhang,
  and Vechev]{dl2}
M.~Fischer, M.~Balunovic, D.~Drachsler-Cohen, T.~Gehr, C.~Zhang, and M.~Vechev.
\newblock Dl2: Training and querying neural networks with logic.
\newblock In \emph{International Conference on Machine Learning}, pages
  1931--1941, 2019.

\bibitem[Guimar{\~{a}}es and Costa(2021)]{NeuralLog}
V.~Guimar{\~{a}}es and V.~S. Costa.
\newblock Neurallog: a neural logic language.
\newblock \emph{CoRR}, abs/2105.01442, 2021.
\newblock URL \url{https://arxiv.org/abs/2105.01442}.

\bibitem[Hinton et~al.(2012)Hinton, Deng, Yu, Dahl, Mohamed, Jaitly, Senior,
  Vanhoucke, Nguyen, Kingsbury, et~al.]{speech_recognition}
G.~Hinton, L.~Deng, D.~Yu, G.~Dahl, A.-r. Mohamed, N.~Jaitly, A.~Senior,
  V.~Vanhoucke, P.~Nguyen, B.~Kingsbury, et~al.
\newblock Deep neural networks for acoustic modeling in speech recognition.
\newblock \emph{IEEE Signal processing magazine}, 29, 2012.

\bibitem[Hoernle et~al.(2021)Hoernle, Karampatsis, Belle, and
  Gal]{MultiplexNet}
N.~Hoernle, R.~Karampatsis, V.~Belle, and K.~Gal.
\newblock Multiplexnet: Towards fully satisfied logical constraints in neural
  networks.
\newblock \emph{CoRR}, abs/2111.01564, 2021.
\newblock URL \url{https://arxiv.org/abs/2111.01564}.

\bibitem[Hu et~al.(2016)Hu, Ma, Liu, Hovy, and Xing]{harnessing}
Z.~Hu, X.~Ma, Z.~Liu, E.~H. Hovy, and E.~P. Xing.
\newblock Harnessing deep neural networks with logic rules.
\newblock In \emph{Proceedings of the 54th Annual Meeting of the Association
  for Computational Linguistics, {ACL} 2016, August 7-12, 2016, Berlin,
  Germany, Volume 1: Long Papers}. The Association for Computer Linguistics,
  2016.
\newblock ISBN 978-1-945626-00-5.
\newblock URL \url{http://aclweb.org/anthology/P/P16/P16-1228.pdf}.

\bibitem[Koller et~al.(2007)Koller, Friedman, D{\v{z}}eroski, Sutton, McCallum,
  Pfeffer, Abbeel, Wong, Heckerman, Meek, et~al.]{introduction_SRL}
D.~Koller, N.~Friedman, S.~D{\v{z}}eroski, C.~Sutton, A.~McCallum, A.~Pfeffer,
  P.~Abbeel, M.-F. Wong, D.~Heckerman, C.~Meek, et~al.
\newblock \emph{Introduction to statistical relational learning}.
\newblock MIT press, 2007.

\bibitem[Krizhevsky et~al.(2012)Krizhevsky, Sutskever, and
  Hinton]{image_classification}
A.~Krizhevsky, I.~Sutskever, and G.~E. Hinton.
\newblock Imagenet classification with deep convolutional neural networks.
\newblock In \emph{Proceedings of the 25th International Conference on Neural
  Information Processing Systems - Volume 1}, NIPS'12, pages 1097--1105, USA,
  2012. Curran Associates Inc.
\newblock URL \url{http://dl.acm.org/citation.cfm?id=2999134.2999257}.

\bibitem[Li and Srikumar(2019)]{augmenting}
T.~Li and V.~Srikumar.
\newblock Augmenting neural networks with first-order logic.
\newblock In \emph{Proceedings of the 57th Annual Meeting of the Association
  for Computational Linguistics}, pages 292--302, Florence, Italy, July 2019.
  Association for Computational Linguistics.
\newblock \doi{10.18653/v1/P19-1028}.
\newblock URL \url{https://www.aclweb.org/anthology/P19-1028}.

\bibitem[Lu et~al.(2016)Lu, Krishna, Bernstein, and Li]{visual2}
C.~Lu, R.~Krishna, M.~S. Bernstein, and F.~Li.
\newblock Visual relationship detection with language priors.
\newblock In \emph{{ECCV} {(1)}}, volume 9905 of \emph{Lecture Notes in
  Computer Science}, pages 852--869. Springer, 2016.

\bibitem[Lu and Getoor(2003)]{citeseer}
Q.~Lu and L.~Getoor.
\newblock Link-based classification.
\newblock In \emph{Proceedings of the Twentieth International Conference on
  International Conference on Machine Learning}, ICML’03, page 496–503.
  AAAI Press, 2003.
\newblock ISBN 1577351894.

\bibitem[Manhaeve et~al.(2018)Manhaeve, Dumancic, Kimmig, Demeester, and
  De~Raedt]{deepproblog}
R.~Manhaeve, S.~Dumancic, A.~Kimmig, T.~Demeester, and L.~De~Raedt.
\newblock Deepproblog: Neural probabilistic logic programming.
\newblock In \emph{Advances in Neural Information Processing Systems}, pages
  3749--3759, 2018.

\bibitem[Marra et~al.(2020)Marra, Diligenti, Giannini, Gori, and Maggini]{RNM}
G.~Marra, M.~Diligenti, F.~Giannini, M.~Gori, and M.~Maggini.
\newblock Relational neural machines.
\newblock \emph{arXiv preprint arXiv:2002.02193}, 2020.

\bibitem[Minervini and Riedel(2018)]{adversarially}
P.~Minervini and S.~Riedel.
\newblock Adversarially regularising neural nli models to integrate logical
  background knowledge.
\newblock \emph{arXiv preprint arXiv:1808.08609}, 2018.

\bibitem[Reimann and Schwung(2019)]{NLRL}
J.~N. Reimann and A.~Schwung.
\newblock Neural logic rule layers.
\newblock \emph{arXiv preprint arXiv:1907.00878}, 2019.

\bibitem[Richardson and Domingos(2006)]{MLN}
M.~Richardson and P.~Domingos.
\newblock Markov logic networks.
\newblock \emph{Mach. Learn.}, 62\penalty0 (1-2):\penalty0 107--136, Feb. 2006.
\newblock ISSN 0885-6125.

\bibitem[Rockt{\"a}schel and Riedel(2016)]{NTP1}
T.~Rockt{\"a}schel and S.~Riedel.
\newblock Learning knowledge base inference with neural theorem provers.
\newblock In \emph{Proceedings of the 5th Workshop on Automated Knowledge Base
  Construction}, pages 45--50, 2016.

\bibitem[Rockt{\"a}schel and Riedel(2017)]{NTP2}
T.~Rockt{\"a}schel and S.~Riedel.
\newblock End-to-end differentiable proving.
\newblock In \emph{Advances in Neural Information Processing Systems}, pages
  3788--3800, 2017.

\bibitem[Sacc{\'a} et~al.(2013)Sacc{\'a}, Diligenti, and Gori]{SBR_collective}
C.~Sacc{\'a}, M.~Diligenti, and M.~Gori.
\newblock Collective classification using semantic based regularization.
\newblock In \emph{2013 12th International Conference on Machine Learning and
  Applications}, volume~1, pages 283--286. IEEE, 2013.

\bibitem[Sen et~al.(2008)Sen, Namata, Bilgic, Getoor, Gallagher, and
  Eliassi-Rad]{collective_classification}
P.~Sen, G.~M. Namata, M.~Bilgic, L.~Getoor, B.~Gallagher, and T.~Eliassi-Rad.
\newblock Collective classification in network data.
\newblock \emph{AI Magazine}, 29\penalty0 (3):\penalty0 93--106, 2008.

\bibitem[Serafini and d'Avila Garcez(2016)]{LTN}
L.~Serafini and A.~S. d'Avila Garcez.
\newblock Logic tensor networks: Deep learning and logical reasoning from data
  and knowledge.
\newblock \emph{CoRR}, abs/1606.04422, 2016.

\bibitem[Van~Krieken et~al.(2019)Van~Krieken, Acar, and
  Van~Harmelen]{semi_supervised}
E.~Van~Krieken, E.~Acar, and F.~Van~Harmelen.
\newblock Semi-supervised learning using differentiable reasoning.
\newblock \emph{arXiv preprint arXiv:1908.04700}, 2019.

\bibitem[Xu et~al.(2018)Xu, Zhang, Friedman, Liang, and Van~den
  Broeck]{semantic_loss}
J.~Xu, Z.~Zhang, T.~Friedman, Y.~Liang, and G.~Van~den Broeck.
\newblock A semantic loss function for deep learning with symbolic knowledge.
\newblock In J.~Dy and A.~Krause, editors, \emph{Proceedings of the 35th
  International Conference on Machine Learning}, volume~80 of \emph{Proceedings
  of Machine Learning Research}, pages 5502--5511, Stockholmsmässan, Stockholm
  Sweden, 10--15 Jul 2018. PMLR.
\newblock URL \url{http://proceedings.mlr.press/v80/xu18h.html}.

\end{thebibliography}
